# Knowledge-guided Text Structuring in Clinical Trials

Yingcheng Sun and Kenneth Loparo

Case Western Reserve University, Cleveland OH 44106, USA
{yxs489,kal4}@case.edu

**Abstract.** Clinical trial records are variable resources or the analysis of patients and diseases. Information extraction from free text such as eligibility criteria and summary of results and conclusions in clinical trials would better support computer-based eligibility query formulation and electronic patient screening. Previous research has focused on extracting information from eligibility criteria, with usually a single pair of medical entity and attribute, but seldom considering other kinds of free text with multiple entities, attributes and relations that are more complex for parsing. In this paper, we propose a knowledge-guided text structuring framework with an automatically generated knowledge base as training corpus and word dependency relations as context information to transfer free text into formal, computer-interpretable representations. Experimental results show that our method can achieve overall high precision and recall, demonstrating the effectiveness and efficiency of the proposed method.

## 1    Introduction

The growing volume of clinical trial records are one of the most valuable sources of evidence on the efficacy of treatments in humans. In clinical trial records, many items like eligibility criteria and summary of results and conclusions are usually written in unstructured free text. The free – text format and lack of standardization have inhibited the effective use for automatic identification of eligible patients from electronic health records (EHRs), and makes extracting the target data for evidence - based synthesis burdensome.

To structure the free text in clinical trial records, three major elements need to be extracted: entity, attribute and the relations between them. Figure 1 shows an example of eligibility criteria with three types of elements.

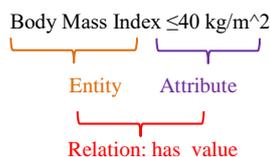

**Fig. 1.** Example of free text annotated by entity, attribute and relation between them

"Body Mass Index" is the medical entity, often identified as Unified Medical Language System (UMLS) concept with an unique identifier (CUI). "≤40 kg/m^2" is the attribute of "Body Mass Index" with value 40, unit "kg/m^2" and comparison "≤", and "has_value" is the relationship between the entity and its corresponding attribute. Most previous work has focused on structuring eligibility criteria that usually only contains a single pair of entity and attribute in one sentence like the example in Figure 1, while information extraction in other types of free text has received little attention. The free text of a clinical trial like summary of results and conclusions may contain more than one medical entity and associated attributes, with multiple mappings needing to be inferred. Figure 2 shows examples of two long paragraphs of free text with multiple entities and attributes.

---

Patients who are taking any concomitant medications that might confound assessments of pain relief, such as psychotropic drugs, antidepressants, sedative hypnotics or any analgesics taken within three days or five times of their elimination half-lives, whichever is longer. Selective serotonin reuptake inhibitors (SSRIs) and selective noradrenaline reuptake inhibitors (SNRIs) are permitted if the patient has been on a stable dose for at least 30 days prior to screening. [NCT04018612]

M/F ages 21-45 with a history of smoked cocaine use at least twice a week for the past six months. A normal resting 12-lead electrocardiograph (ECG) and blood pressure of less than 140/90 mmHg. [NCT01640873]

---

**Fig. 2.** Examples of long paragraphs with multiple entities and attributes. They is acquired from ClinicalTrials.gov

The possible entities in the first paragraph are: medications, pain, psychotropic drugs, antidepressants, sedative hypnotics, analgesics, elimination half-lives, screening, Selective serotonin reuptake inhibitors (SSRIs) and selective noradrenaline reuptake inhibitors (SNRIs). The possible attributes are concomitant, within three days, five times, at least 30 days prior. After comparing all the information extraction tools, especially those for eligibility criteria such as cTAKES, MetaMap and Criteria2query, we find that there are no tools so far that can extract all the entities and attributes correctly, and neither do their relations after our experiments. For the second paragraph, we found the same issues: entity and attributes are missing, and relations extracted are not accurate. Therefore, extracting all the possible candidates and detecting the correct relations among all candidates becomes an important concern.

Because of the complexities of inferring "many-to-many" relationships, simple text parsing methods, such as regular expressions, may not be an effective way to extract and structure the elements in the target text [15, 19]. Machine learning models are adopted in some systems for the relation extraction task, such as support vector machine and convolutional neural networks [23]. However, traditional machine learning models rely on human annotation for training data generation, which is time and labor consuming. Moreover, the reliance on human annotation further limits the systems to certain pre-specified relational types and makes it unable to further extend the systems to new relational types.

Each medical entity has its own value range and unit format that can be easily obtained from an online knowledge base such as Wikipedia and Observational Health Data Sciences and Informatics (OHDSI), so those formats can be used as attribute constraints when inferring the relations [13, 14, 16, 17]. In this paper, we present a novel distant supervision approach that generates training data automatically by aligning texts and a knowledge base, and learns to automatically extract relations between a medical entity and its corresponding attribute. We also use the syntactical structure of a sentence to assign the mapping direction because all relations are directional, pointing from the entity to its attribute. Experimental results show that our method achieves overall high precision and recall, demonstrating the effectiveness and efficiency of the proposed method.

The rest of the paper is organized as follows. In Section 2, we present related work. In Section 3, we propose the knowledge-based distant supervision approach. In Section 4, we introduce the datasets, comparison methods and evaluation metrics, as well as experimental results. We conclude our work in Section 5.

## 2 Related Work

Parsing clinical text is important to leverage the reuse of clinical information for automatic decision support. Following this assumption, various methods and techniques have been recently developed to transform clinical trial protocol text into computable representations that can be used to facilitate automated tasks.

### 2.1 Ontology-based and rule-based approaches

Riccardo et al. [1] proposed a clinical trial search framework eTACTS that combines tag mining of free-text eligibility criteria and interactive retrieval based on dynamic tag clouds to reduce the number of resulting trials returned by a simple search. Tu et al. [2] designed the Eligibility Rule Grammar and Ontology for clinical eligibility criteria and used it to transform free-text eligibility criteria into computable criteria [3]. Bhattacharya and Cantor [4] proposed a template-based representation of eligibility criteria for standardizing criteria statements. Weng et al. [5] leveraged text mining to develop a Unified Medical Language System – a semantic like network-based representation for clinical trial eligibility criteria called EliXR.

### 2.2 Machine learning based approaches

Methods based on machine learning models for standardizing categorical eligibility criteria were developed, such as EliXRTIME [6] for temporal knowledge representation and Valx [7] for numeric expression extraction and normalization. Much of the more recent work on distant supervision since has been focused on the task of relation extraction [8, 9] and classification of Twitter/microblog texts [10]. Supervised distant supervision is used to automatically extract Population/Problem, Intervention, Comparator, and Outcome (PICO criteria) in clinical trial records [11].

Despite these efforts, there is still a need to develop a solution for directly transforming all kinds of free text in clinical trial records to structured information, with a computational output data format. We explored the method of information extraction from free text in clinical trials with knowledge-based distant supervision in our previous research [18, 20, 21, 22], and provide more details in this paper.

## 3    Knowledge-Based Distant Supervision Framework

In this section, we propose a probabilistic linking model to deal with the task of entity linking with its corresponding attribute. The overall framework is shown in Fig. 2.

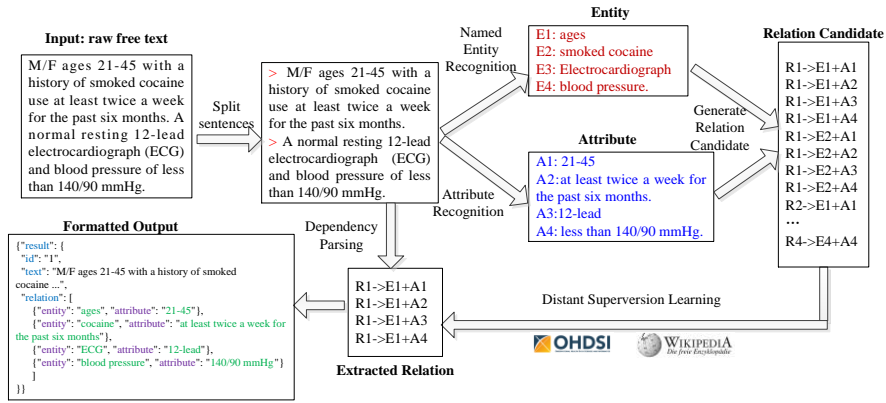

**Fig. 3.** The overview of the knowledge-based distant supervision framework

The inputs are text paragraphs from a clinical trial corpus. Generally, eligibility criteria are organized by lines, but other types of clinical records are written in paragraphs, so the first step is to split the input text into sentences. After separating the input text into sentences, medical entities need to be annotated as UMLS concept and attributes need to be recognized as temporal expressions or value expressions. We use part of speech(POS) pattern mining to build the knowledge base from the knowledge sources medical dictionary, Wikipedia and OHDSI as Fig. 4 shows.

We use a small training dataset to train the random forest classifier, and refine the results manually. In the knowledge base table, the first column is "term", which lists medical concepts. They can be used to improve the entity recognition. The second and third columns are value and units that can be used to improve the relation extractions. For example, suppose we identified two concepts: "blood pressure" and "body weights", and have two values as their attributes. It is highly possible that the value ending with "mmHg" will be blood pressure's attribute and the value ending with "kg" will be body weight's attribute because they are their usually used units. We found that unit is more useful than value in the relations recognition because in clinical trials, the values from patients are usually not in a normal range.

**Knowledge Base**

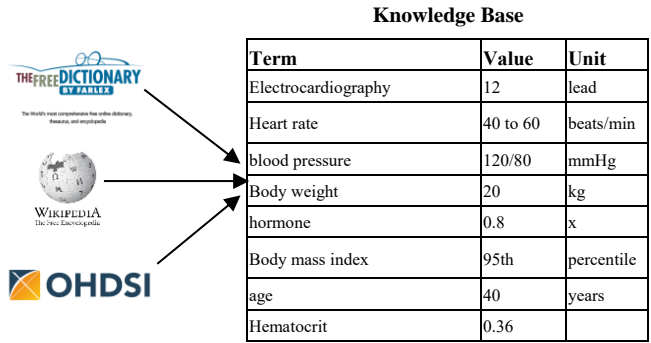

| Term | Value | Unit |
|------|-------|------|
| Electrocardiography | 12 | lead |
| Heart rate | 40 to 60 | beats/min |
| blood pressure | 120/80 | mmHg |
| Body weight | 20 | kg |
| hormone | 0.8 | x |
| Body mass index | 95th | percentile |
| age | 40 | years |
| Hematocrit | 0.36 | |

**Fig. 4.** Knowledge sources and a snippet of the knowledge base

We updated the algorithms in Criteria2Query by introducing the knowledge base for named entity recognition [12]. After this step, M entities and N attributes are extracted from the free-text record automatically, and then we have M*N relational candidates. Next, we use the Stanford dependency parser CoreNLP to discover the syntactical structure of a sentence.

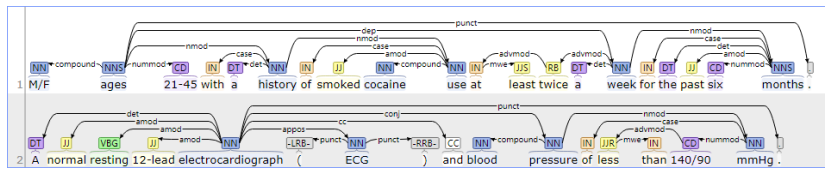

**Fig. 5.** Word dependency extracted by CoreNLP for of the second paragraph in Fig.1

But this tool will extract all the relations among all the words. Most of them are not our targets. So we simplified the CoreNLP tool and only extracted relations that we concern based on identified named entities in last step, as Fig. 6 shows.

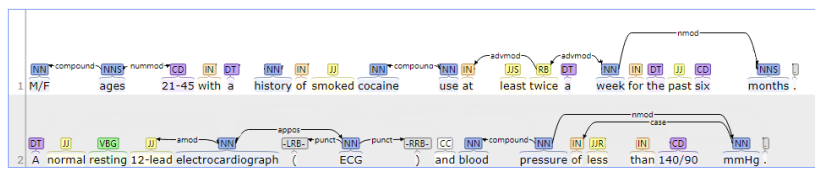

**Fig. 6.** The simplified word dependency extraction result for the second paragraph in Fig.1

We assume that there is an underlying distribution P over the relations given the parsing results, and then the probability of a relation r whose dependency parsing result is dep could be expressed using the following formula:

$$P(r|dep) = P(e, a| dep)$$

Where $e$ is the entity and $a$ is the attribute. Meanwhile, distant supervision is applied on all candidate relations and the probability of a relation $r$ whose distant supervision learning result is $sup$ could be expressed as the following formula:

$$P(r|sup) = P(e, a| sup)$$

We use Wikipedia and OHDSI as the knowledge base, and model the entity by its value range and unit. For example, the standard written format for blood pressure is" $\backslash d+/\backslash d$", and its unit is "mmHg", so "115/75 mmHg" would have higher probability to be a blood pressure than "11-25". Given the two types of relational probabilities, we could further define the relational assignment probability as:

$$P(r|s)= \theta \cdot P(r|sup) + (1-\theta) \cdot P(r|dep)$$

Where $\theta$ is a parameter that balances the two parts and $s$ is the sentence. Next, we select the probability with largest value i.e., the most likely mapping between the entity $e$ and attribute $a$ given a sentence s as context.

After the recognition and extraction of entities and their attributes and identification of the relations between them, the entities were encoded by the standard UMLS terminology. To support downstream large-scale analysis, we standardized the output results in the JSON format. An example output format is illustrated in Figure 2.

## 4    Experiment

To evaluate the efficiency of our proposed framework, we assess the accuracy and recall of the information extraction in this section. The experimental setup includes a ThinkPad laptop with Intel Core i5 2.50 GHz processor and 10 GB RAM. In our experiment, we use the clinical trial data obtained from ClinicalTrials.gov. 100 clinical trial records were randomly picked and 477 free-text records including eligibility criteria and summary of results and conclusions were acquired. Next, three annotators were asked to label each record with entity, attribute and their relations, and we used the labels agreed to by at least two annotators as the ground truth, with a total of 386 records labeled in this manner. All annotations were completed in Brat [24], a web-based annotation tool. Fig. 7 shows the results.

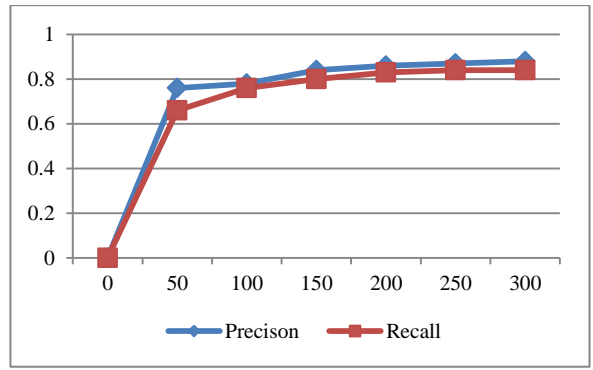

**Fig. 7.** Learning curves for recognition tasks by different sizes of training sizes.

The results show that when the number of training data is over 200, the performance reaches a stable status, and the average accuracy is 0.83, and the average recall is 0.79 that show relatively high effectiveness. Fig. 8 shows an example of the JSON format output. We do not parse the attribute further because we focus more on the issue of relationship inference. Several tools can be used to separate the attributes into fine-grained parts.

```
{"result": {
    "id": "1",
    "text": " M/F ages 21-45 with a history of smoked cocaine use at least twice a week for
the past six months. A normal resting 12-lead electrocardiograph (ECG) and blood pressure of
less than 140/90 mmHg.",
    "relation": [
        {"entity": "ages", "attribute": "21-45"},
        {"entity": "cocaine", "attribute": "at least twice a week for the past six months"},
        {"entity": "ECG", "attribute": "12-lead"},
        {"entity": "blood pressure", "attribute": "140/90 mmHg"}
        ]
}}
```

**Fig. 8.** Example of the JSON format output

## 5 Discussion And Future Work

Structured clinical trial records can enable more rigorous comparisons of trial populations and better meta-analyses of collective generalizability of related clinical trials and also pave the path to thorough studies of clinical trial participants and their representativeness of the general population. In this paper, we proposed a knowledge - based distant supervision framework to extract information from all types of free text data in clinical trials records. Our experiments demonstrate the effectiveness of our method. In the future, we will explore the active learning methods [18] or reinforcement learning [20, 22] to annotate the criteria eligibility to reduce the annotation cost. Also, we are interested in constructing the query expression for clinical trials records and build the interface to search all kinds of clinical trial records.

## Acknowledgment

This work was supported by the Ohio Department of Higher Education, the Ohio Federal Research Network and the Wright State Applied Research Corporation under award WSARC-16-00530 (C4ISR: Human-Centered Big Data).